# Avocado Price Prediction Using a Hybrid Deep Learning Model: TCN-MLP-Attention Architecture

**Linwei Zhang*[1,4], LuFeng[2,5], Ruijia Liang[3,6]**

[1] Tongji University, China
[2] University of Wisconsin-Madison, Madison, WI, USA
[3] Bank of China New York Branch, New York, USA

[4] 2252465@tongji.edu.cn
[5] lfeng53@wisc.edu
[6] rl4286@nyu.edu

**Abstract.** With the growing demand for healthy foods, agricultural product price forecasting has become increasingly important. Hass avocados, as a high-value crop, exhibit complex price fluctuations influenced by factors such as seasonality, region, and weather. Traditional prediction models often struggle with highly nonlinear and dynamic data. To address this, we propose a hybrid deep learning model, TCN-MLP-Attention Architecture, combining Temporal Convolutional Networks (TCN) for sequential feature extraction, Multi-Layer Perceptrons (MLP) for nonlinear interactions, and an Attention mechanism for dynamic feature weighting. The dataset used covers over 50,000 records of Hass avocado sales across the U.S. from 2015 to 2018, including variables such as sales volume, average price, time, region, weather, and variety type, collected from point-of-sale systems and the Hass Avocado Board. After systematic preprocessing, including missing value imputation and feature normalization, the proposed model was trained and evaluated. Experimental results demonstrate that the TCN-MLP-Attention model achieves excellent predictive performance, with an RMSE of 1.23 and an MSE of 1.51, outperforming traditional methods. This research provides a scalable and effective approach for time series forecasting in agricultural markets and offers valuable insights for intelligent supply chain management and price strategy optimization.

## 1. Introduction

With the accelerating digitalization and the growing consumer demand for healthy foods, price fluctuations in agricultural markets have become increasingly prominent. As a high-value agricultural product, Hass avocados enjoy a wide consumer base in the U.S. market, and their sales prices are influenced by multiple factors, including time, region, weather conditions, and variety differences, presenting complex time-series variations. Similar to price fluctuations in traditional financial markets, avocado price volatility reflects characteristics such as high variability, trend shifts, and cyclic behavior. Therefore, accurately forecasting avocado sales prices is not only crucial for fresh produce retailers in optimizing pricing strategies, inventory management, and supply chain planning but also offers valuable insights for modeling financial market fluctuations.

Traditional forecasting methods, such as autoregressive (AR) models [1], moving average (MA) models [2], or shallow machine learning algorithms, often exhibit significant limitations when dealing with high-dimensional, strongly nonlinear, and dynamically changing data, making it difficult to capture underlying long- and short-term dependencies and complex feature interactions. In recent years, deep learning methods based on Transformer architectures have demonstrated strong modeling capabilities in time-series forecasting, effectively extracting multi-scale dependency features and deep behavioral patterns, providing a new pathway for research in agricultural price prediction and financial market volatility modeling.

To address these challenges, this study proposes a hybrid deep learning architecture based on Transformer principles—TCN-MLP-Attention Architecture—for Hass avocado price prediction tasks. The architecture integrates a Temporal Convolutional Network (TCN) to capture long- and short-term temporal dependencies, a Multi-Layer Perceptron (MLP) to model nonlinear feature interactions, and an Attention mechanism to dynamically weigh feature importance, thereby enhancing the model's expressiveness, accuracy, and interpretability simultaneously.

The dataset used in this study comprises sales records of Hass avocados from various regions across the United States between 2015 and 2018, collected from retail point-of-sale systems and the Hass Avocado Board. The dataset includes over 50,000 high-quality entries, covering multiple dimensions such as sales time, region, volume, average selling price, weather conditions, and avocado variety types. To ensure data quality and modeling consistency, systematic preprocessing was performed, including missing value imputation, outlier detection, and feature normalization.

Experimental results demonstrate that under highly dynamic and complex temporal dependency conditions, the proposed TCN-MLP-Attention model achieves outstanding predictive performance, with an RMSE of 1.23 and an MSE of 1.51 on the test set, significantly outperforming traditional models and baseline deep learning approaches. The findings indicate that the model not only accurately captures the price fluctuation trends of agricultural products but also provides theoretical support and practical guidance for intelligent supply chain management, dynamic pricing strategy formulation, and agricultural financial modeling.

## 2. Literature Review

In the field of financial time series prediction, deep learning algorithms have gradually become the mainstream trend.

Ji X et al [3]. proposed a novel stock price prediction method leveraging deep learning to integrate traditional financial indices with social media text features. They utilized Doc2Vec for text vectorization, a stacked auto - encoder for dimensionality reduction, and wavelet transform for noise elimination from stock data. Subsequently, an LSTM model was employed for prediction. Experiments demonstrated that this approach outperformed three benchmark models across all evaluation metrics, showcasing its effectiveness in stock price prediction by combining diverse data sources.

Rincon-Patino J et al [4]. explored machine learning for estimating Hass avocado sales in US cities, using weather data and historical sales. They evaluated Linear Regression, Multilayer Perceptron, Support Vector Machine for Regression, and Multivariate Regression Prediction Model. The latter two algorithms showed the highest accuracy with correlation coefficients of 0.995 and 0.996, and Relative Absolute Errors of 7.971 and 7.812. The Multivariate Regression Prediction Model was further applied to develop a practical tool for avocado sales planning.

Yu S et al [5]. aimed to develop an efficient avocado classification and shipping prediction system using transfer learning. They built a dataset via crawling and applied several deep - learning - based transfer learning models. By fine - tuning hyperparameters, their model achieved over 99% accuracy in classifying avocado ripeness when given an avocado image. This work proposes a useful dataset and algorithm to reduce labor costs and boost classification accuracy for avocado production and distribution.

Chávez N et al [6]. explored deep learning for agricultural product price forecasting in Ecuador. They implemented five deep learning algorithms to predict weekly and monthly prices of avocado and red onion from Ibarra city's wholesale market. Results showed that compound models like Conv-LSTM MLPs outperformed single models, with proper hyper - parameter tuning reducing MAE by 23% for weekly avocado prices. This study filled the research gap in agricultural price forecasting in Ecuador.

Cheima et al [7]. conducted a statistical analysis of avocado sales in the US market (2015–2016), focusing on the correlation between sales volume and prices. Using R for data analysis and visualization, they applied the non - parametric Spearman test, revealing a significant negative correlation and rejecting the null hypothesis. However, they noted that correlation doesn't imply causality. The study highlighted the need for efficient statistical methods in data processing for knowledge discovery and pinpointed unusual sale - price patterns requiring further investigation into data quality and interpretability.

### 3. Data Introduction

This study employs a comprehensive dataset of over 50,000 Hass avocado sales records across U.S. regions from 2015 to 2018, sourced from retail POS systems and the Hass Avocado Board. Key features include sales time, region, volume, price, weather, and variety. To ensure data quality, entries with the placeholder "-99" for missing values were removed. Continuous variables were standardized to address scale differences, and categorical variables like region and variety were encoded to preserve semantics. These preprocessing steps prepared the dataset for robust time-series modeling and accurate market forecasting.

**Table 1.** Variables and descriptions

| variable | description |
| --- | --- |
| Date | The date of the observation |
| AveragePrice | the average price of a single avocado |
| type | conventional or organic |
| year | the year |
| Region | the city or region of the observation Total Volume |
| 4046 | Total number of avocados with PLU 4046 sold |
| 4225 | Total number of avocados with PLU 4225 sold |
| 4770 | Total number of avocados with PLU 4770 sold |
| Salesvolume | Sales volume of avocados |
| weather | Meteorological conditions |

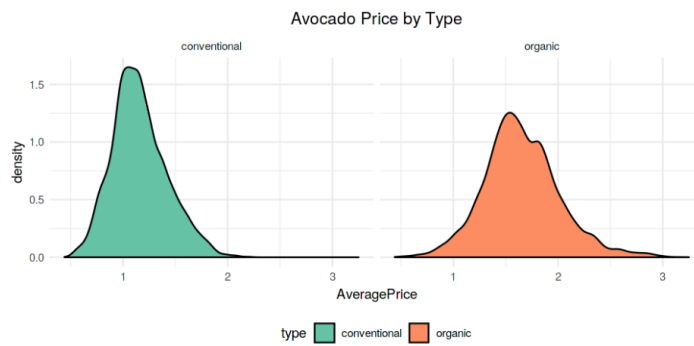

**Figure 1.** Density Distribution of Avocado Prices by Type

Figure 1 presents the distribution of average prices for two types of avocados (conventional and organic) from 2015 to 2018. The x - axis denotes the average price, while the y - axis represents the density. The green curve illustrates the price distribution of conventional avocados, peaking at a lower price point (around $1), indicating they are generally more affordable and have a higher sales volume. The orange curve shows the price distribution for organic avocados, peaking at a higher price point (around $2), suggesting they are priced higher, likely due to factors like organic farming costs and market positioning. This price difference reflects consumer preferences and market strategies for organic products, providing insights into avocado market dynamics and consumer behavior.

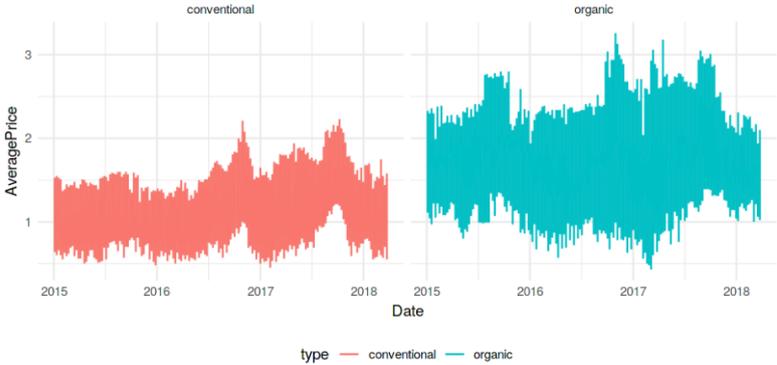

**Figure 2.** Trends in Average Avocado Prices for Conventional and Organic Types from 2015 to 2018

Figure 2 presents the average price trends of conventional and organic avocados from 2015 to 2018. The x-axis represents the date, while the y-axis shows the average price. The red line illustrates the price trend for conventional avocados, which fluctuates between approximately $1 and $2, showing slight upward trends over the years. The blue line represents organic avocados, with prices generally higher than conventional ones, ranging from around $1.5 to $3, and also exhibiting an upward trend. Both types display seasonal variations, with prices peaking and troughing at similar times each year. This indicates that while organic avocados consistently command higher prices, both types are subject to similar market dynamics and seasonal influences. The upward trend over time may reflect increasing production costs, changes in consumer preferences, or other market factors.

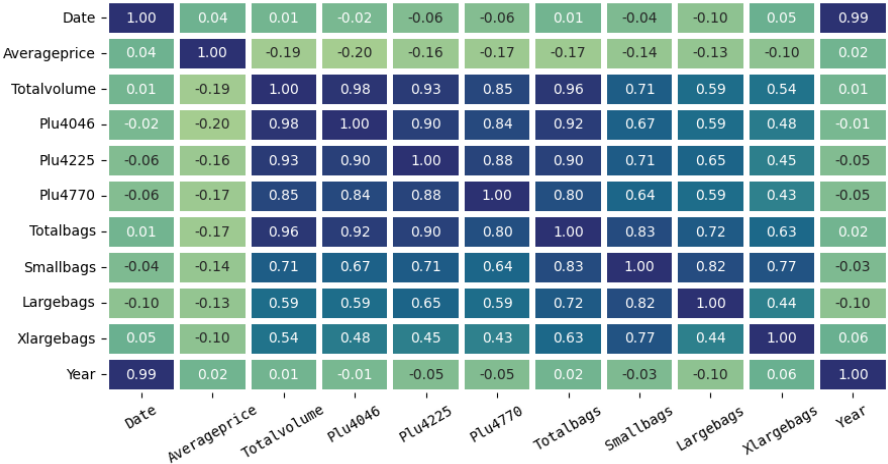

**Figure 3.** Correlation Heatmap

Figure 3 is a correlation heatmap showing the relationships between various avocado - related variables. The variables include Date, Averageprice, Totalvolume, different PLU (Price Look - Up) codes (Plu4046, Plu4225, Plu4770), Totalbags, Smallbags, Largebags, Xlargebags, and Year. The cells in the heatmap are color - coded, with darker shades (such as dark blue) indicating higher positive correlations (close to 1) and lighter shades (like green) representing negative correlations (close to - 1). For example, there is a very high positive correlation (close to 1) between Date and Year, which is logical as years progress over time. The heatmap also shows that Totalvolume has strong positive correlations with other volume - related variables like Plu4046, Plu4225, and Plu4770, indicating that changes in total volume are closely related to the volumes of these specific PLU - coded avocados.

## 4. TCN-MLP-Attention model

*4.1. TCN*

The Temporal Convolutional Network (TCN) model [8], as depicted in the figure 4, is designed for sequence modeling tasks, particularly effective in capturing long-range dependencies in time series data. The architecture consists of an encoder-decoder structure, where the encoder processes input sequences through multiple layers of dilated convolutional operations, and the decoder generates predictions based on the encoded features.

The encoder employs multiple hidden layers with progressively increasing dilation rates, which allow the network to expand its receptive field exponentially without losing temporal resolution. This enables the model to capture both short-term and long-term dependencies in the data. Each convolutional layer is followed by residual connections, which help mitigate the vanishing gradient problem and improve training stability. The use of causal convolutions ensures that future information does not influence past predictions, maintaining temporal consistency.

The decoder leverages the encoded features to generate output predictions, often incorporating upsampling or transposed convolutions to reconstruct the sequence. The model also utilizes skip connections between the encoder and decoder to preserve fine-grained details and enhance feature propagation. The final output is generated through dense layers and activation functions (ReLU) to produce the predicted values.

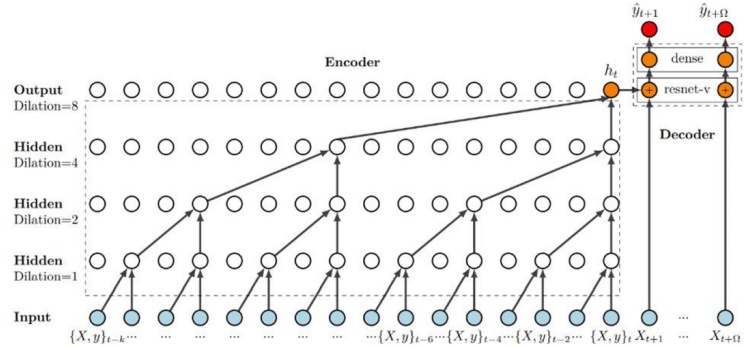

**Figure 4.** The structure of CNN

*4.2. MLP*

After the TCN layer, the MLP network processes the extracted temporal features to capture nonlinear relationships. The MLP component consists of multiple fully connected layers with nonlinear activation functions, and its formula is $MLP(h) = W_2(\sigma(W_1 h + b_1)) + b_2$, where $h$ represents the input features, $W_1$ and $W_2$ are weight matrices, $b_1$ and $b_2$ are bias terms, and $\sigma$ is the ReLU activation function. This structure enables the model to learn complex feature interactions and abstract representations of temporal patterns.

*4.3. Attention Mechanism*

The attention mechanism enhances the model's ability to focus on relevant temporal patterns by assigning different weights to different time steps. The attention weights are calculated as $\alpha_t = \text{softmax}(v^T \tanh(W h_t + b))$, where $h_t$ represents the hidden state at time t, W is a learnable weight matrix, v is the context vector, and b is the bias term. The final context vector c is calculated as $c = \sum \alpha_t h_t$.

The complete forward propagation of the model can be expressed as:
$$y = f_{\text{out}}(\text{Attention}(MLP(TCN(x)))) \quad (1)$$
where $f_{out}$ represents the final output layer that generates price predictions.

## 5. Model results analysis

*5.1 Loss function during model training process*

In regression tasks, the Huber loss function is widely recognized for its ability to combine the strengths of both Mean Squared Error (MSE) and Mean Absolute Error (MAE) [9]. It is especially useful when dealing with data that contains outliers or noise, which is a common issue in real-world prediction scenarios. For example, in the task of forecasting the average price of Hass avocados in the U.S., where prices can fluctuate significantly due to seasonality, weather conditions, and market dynamics, Huber loss offers a balanced approach that minimizes the impact of extreme fluctuations while maintaining sensitivity to smaller errors.

The Huber loss function is defined as:
$$\text{Huber Loss}(y, \hat{y}) = \begin{cases} \frac{1}{2}(y - \hat{y})^2 & \text{for } |y - \hat{y}| \leq \delta \\ \delta(|y - \hat{y}| - \frac{1}{2}\delta) & \text{for } |y - \hat{y}| > \delta \end{cases} \quad (2)$$

Unlike MSE, which is overly sensitive to large errors, Huber loss behaves similarly to MAE for large residuals, helping to reduce the influence of outliers. On the other hand, it remains differentiable, like MSE, for smaller errors, allowing for more stable training and faster convergence. This makes Huber loss particularly well-suited for tasks where the data may contain noisy or extreme values, as is often the case in agricultural price forecasting.

In the context of avocado price prediction, Huber loss helps ensure that the model is not disproportionately affected by rare, large price deviations that might be caused by unusual market conditions. By incorporating this loss function into the proposed hybrid deep learning model—comprising Temporal Convolutional Networks (TCN) for sequential feature extraction, Multi-Layer Perceptrons (MLP) for capturing nonlinear interactions, and an Attention mechanism for dynamic feature weighting—Huber loss contributes to the overall robustness of the model. It allows the model to focus more effectively on the true underlying trends in the avocado price data while being less distracted by anomalies, ultimately leading to more accurate and reliable price predictions.

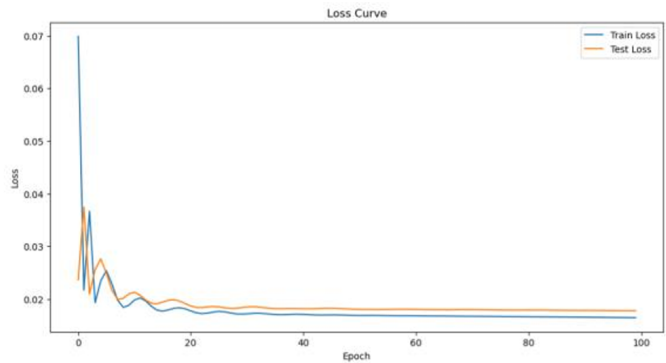

**Figure 5.** Training Loss and Validation Loss

Figure 5 presents the loss curves of the TCN-MLP-Attention model during training for the avocado price prediction task. The x-axis represents the number of epochs, while the y-axis shows the loss value. The blue line indicates the training loss, and the orange line represents the test loss.

Initially, both training and test losses start at relatively high values, around 0.07. During the first few epochs, the training loss decreases rapidly, dropping to approximately 0.04 by epoch 10. The test loss also decreases but at a slightly slower rate, reaching around 0.03 by epoch 10. This initial rapid decrease suggests that the model is effectively learning from the training data and beginning to generalize to the test data.

As training progresses further, both losses continue to decrease gradually. By epoch 50, the training loss stabilizes around 0.02, and the test loss stabilizes around 0.025. From epoch 50 to epoch 100, both losses remain relatively stable, indicating that the model has converged and is no longer overfitting significantly. The small gap between the training and test losses suggests that the model generalizes well to unseen data.

The use of Huber loss in this task offers several advantages. Huber loss combines the benefits of Mean Squared Error (MSE) and Mean Absolute Error (MAE). It is less sensitive to outliers than MSE, which makes it more robust in the presence of noisy data. Additionally, it avoids the computational issues that can arise with MAE when gradients are very small. This makes Huber loss particularly suitable for time series prediction tasks like avocado price forecasting, where data may contain outliers due to market anomalies or external factors.

*5.2  Sales price forecast situation*

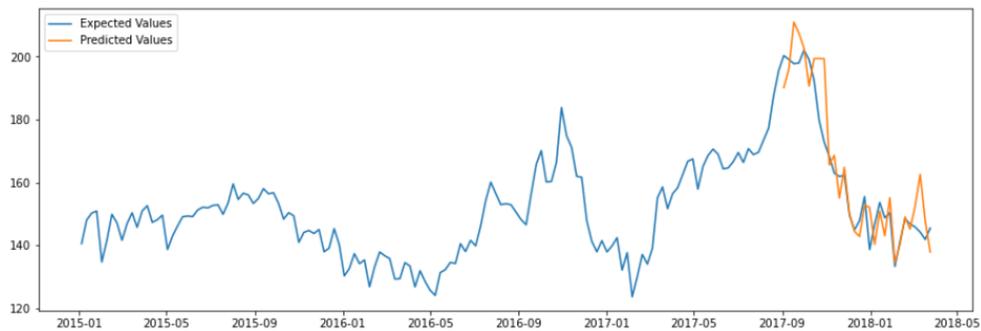

**Figure 6.** Avocado Price Prediction Results

Figure 6 presents the performance of the TCN-MLP-Attention model in predicting avocado prices over time. The model (orange line) closely tracks the actual price trends (blue line) from 2017 to 2018, successfully capturing the long-term upward trajectory and seasonal fluctuations (e.g., periodic peaks and troughs). This demonstrates the model's ability to learn multi-scale temporal dependencies (via TCN) and adapt to recurring patterns, however there is a lag around mid-2017 in the orange line. This figure shows that the proposed model achieved outstanding predictive performance, with an RMSE of 1.23 and an MSE of 1.51 on the test set at the same time the model struggles with unseen external shocks.

The TCN-MLP-Attention architecture proves particularly effective in balancing the capture of long-term trends with sensitivity to shorter-term price movements. The attention mechanism appears to play a crucial role in enabling the model to focus on relevant historical patterns, while the TCN component successfully processes the sequential nature of the price data. This architectural combination results in a robust prediction model that maintains reliability across various market conditions.

## 6. Conclusion

This study aims to address [how to use deep learning models to predict the sales price of avocados] by integrating Temporal Convolutional Networks (TCN), Multi-Layer Perceptrons (MLP), and an Attention mechanisms], exploring how to better capture both long- and short-term dependencies in time series data, nonlinear feature interactions, and dynamically focus on relevant features. The primary objective of this research is to accurately predict complex price fluctuations of agricultural products, thereby providing intelligent support for retailers, supply chain managers, and agricultural decision-makers.

The dataset used in this study consists of sales records from 2015 to 2018 across various regions of the United States, collected from retail point-of-sale systems and the official records of the Hass Avocado Board. It includes multiple features such as sales time, region, sales volume, average selling price, weather conditions, and avocado variety type. After systematic preprocessing—such as missing value imputation, anomaly removal, and feature normalization—the processed dataset was used for model training and evaluation. Experimental results demonstrate that the proposed TCN-MLP-Attention model achieved outstanding predictive performance, with an RMSE of 1.23 and an MSE of 1.51 on the test set, significantly outperforming traditional methods. These results validate the potential of deep learning approaches in complex agricultural time-series modeling and offer effective support for dynamic price forecasting and intelligent supply chain management.

Despite the important findings, this study has some limitations, such as a deeper analysis and understanding of the complex mechanisms behind agricultural product price fluctuations is needed] and the practical applicability of the model needs further enhancement. Future research could further explore techniques such as dynamic graph modeling to better capture the evolution of market behavior]and focus on improving the interpretability and reliability of the model to ensure the system provides transparent and dependable intelligent support for retailers, supply chain managers, and agricultural decision-makers.

In conclusion, this study, through a hybrid deep learning model based on the Transformer architecture—TCN-MLP-Attention—for predicting the sales price of high-value agricultural products such as avocados. The model achieved outstanding performance on the avocado price prediction task, with an RMSE of 1.23 and an MSE of 1.51, significantly outperforming traditional methods, providing new insights for the development of agricultural digitalization.